\documentclass[sigconf,nonacm]{acmart}%
\AtBeginDocument{%
  }

\setcopyright{none}
\renewcommand\footnotetextcopyrightpermission[1]{}
\usepackage{xspace}
\usepackage{array}
\usepackage{booktabs}
\usepackage{makecell}
\usepackage{xcolor}
\usepackage{colortbl}
\usepackage{tabularx}
\usepackage{multirow}
\usepackage{algorithm}
\usepackage{algpseudocode}
\definecolor{PublicOurs}{HTML}{EEF3F8}

\definecolor{PublicGroup}{gray}{0.35}

\makeatletter
\renewcommand{\@mkauthors@iii}{%
  \global\setbox\mktitle@bx=\vbox{%
    \unvbox\mktitle@bx
    \hsize=\textwidth
    \centering
    {\normalfont\large
      Ze Zhao\textsuperscript{1,\textdagger},
      Haoyu Jiang\textsuperscript{1},
      Weiyu Han\textsuperscript{1},
      Ruobing Zuo\textsuperscript{1,2,\textdagger}\\[3pt]
      Xiao Tan\textsuperscript{1,*},
      Shuai Chen\textsuperscript{1},
      Tianyi Zhang\textsuperscript{1}\par}
    \vskip 6pt
    {\normalfont\normalsize
      \textsuperscript{1}Ant International\qquad
      \textsuperscript{2}East China Normal University\par}
    \bigskip
  }%
}
\def\@authornotes{%
  \footnotetext[1]{Corresponding author: \href{mailto:alex.tx@ant-intl.com}{alex.tx@ant-intl.com}.}%
  \footnotetext[2]{Work done during internships at Ant International.}%
}
\makeatother

\begin{document}

\title{ID2Pattern: Pretraining Generalizable Identity Behavior Patterns for Transaction Risk Foundation Models}

\author{Ze Zhao}
\affiliation{\institution{Ant International}\country{}}
\author{Haoyu Jiang}
\affiliation{\institution{Ant International}\country{}}
\author{Weiyu Han}
\affiliation{\institution{Ant International}\country{}}
\author{Ruobing Zuo}
\affiliation{\institution{Ant International}\country{}}
\affiliation{\institution{East China Normal University}\country{}}
\author{Xiao Tan}
\email{alex.tx@ant-intl.com}
\affiliation{\institution{Ant International}\country{}}
\author{Shuai Chen}
\affiliation{\institution{Ant International}\country{}}
\author{Tianyi Zhang}
\affiliation{\institution{Ant International}\country{}}
\renewcommand{\shortauthors}{Zhao et al.}

\newcommand{\method}{\textsc{ID2Pattern}\xspace}
\newcommand{\para}[1]{\smallskip\noindent\textbf{#1.}\ }

\begin{abstract}
Large-scale behavioral pretraining offers a promising route to generalizable representations in online payment systems.
However, high-cardinality identity fields such as card, device, and account IDs, though rich in behavioral signals, remain underexploited.
Their continually growing vocabularies and limited semantic meaning make existing vocabulary-based and semantics-based encodings ill-suited.
Their generalizable signal lies not in specific values, but in the identity behavior patterns they exhibit across a transaction sequence.
Building on this insight, we propose \method, a pretraining framework that brings these patterns into transaction foundation models.
\method first encodes identity fields with a Local Anonymous ID (LAID) tokenizer, which captures how each identity appears and recurs within the sequence, independent of its specific value.
It then pretrains a multi-scale pattern codebook on unlabeled sequences, which abstracts LAID sequences into discrete behavior patterns for downstream prediction.
Experiments on an industrial dataset with over 100M sequences and six public datasets show that \method consistently outperforms strong baselines, improving industrial by 4.13\% and public datasets by 8.83\% on average.
Notably, \method remains equally effective even when all identity values are unseen during training, highlighting its potential to generalize to new users and markets.
\end{abstract}
\keywords{Behavioral Pretraining, Transaction Foundation Models, Fraud Detection, Online Payment}

\maketitle

\section{Introduction}

Behavioral pretraining offers a promising route to transaction foundation models by learning generalizable representations from massive interaction histories.
As online payment becomes core infrastructure of the Web economy, payment and e-commerce platforms on the Web continuously accumulate longitudinal interactions among users, payment instruments, merchants, and other entities, providing rich behavioral signals beyond individual transactions~\cite{jurgovsky2018sequence,padhi2021tabular}.
Recent studies have therefore pretrained behavioral models on large-scale event sequences and adapted them to downstream tasks such as fraud detection and risk assessment~\cite{babaev2022coles,padhi2021tabular,li2025panther}.
These models characterize each transaction event by \textbf{semantic attributes} with relatively stable vocabularies, such as payment types and regions, and learn temporal dependencies across events.

\begin{figure}[h]
    \centering
    \includegraphics[width=\linewidth]{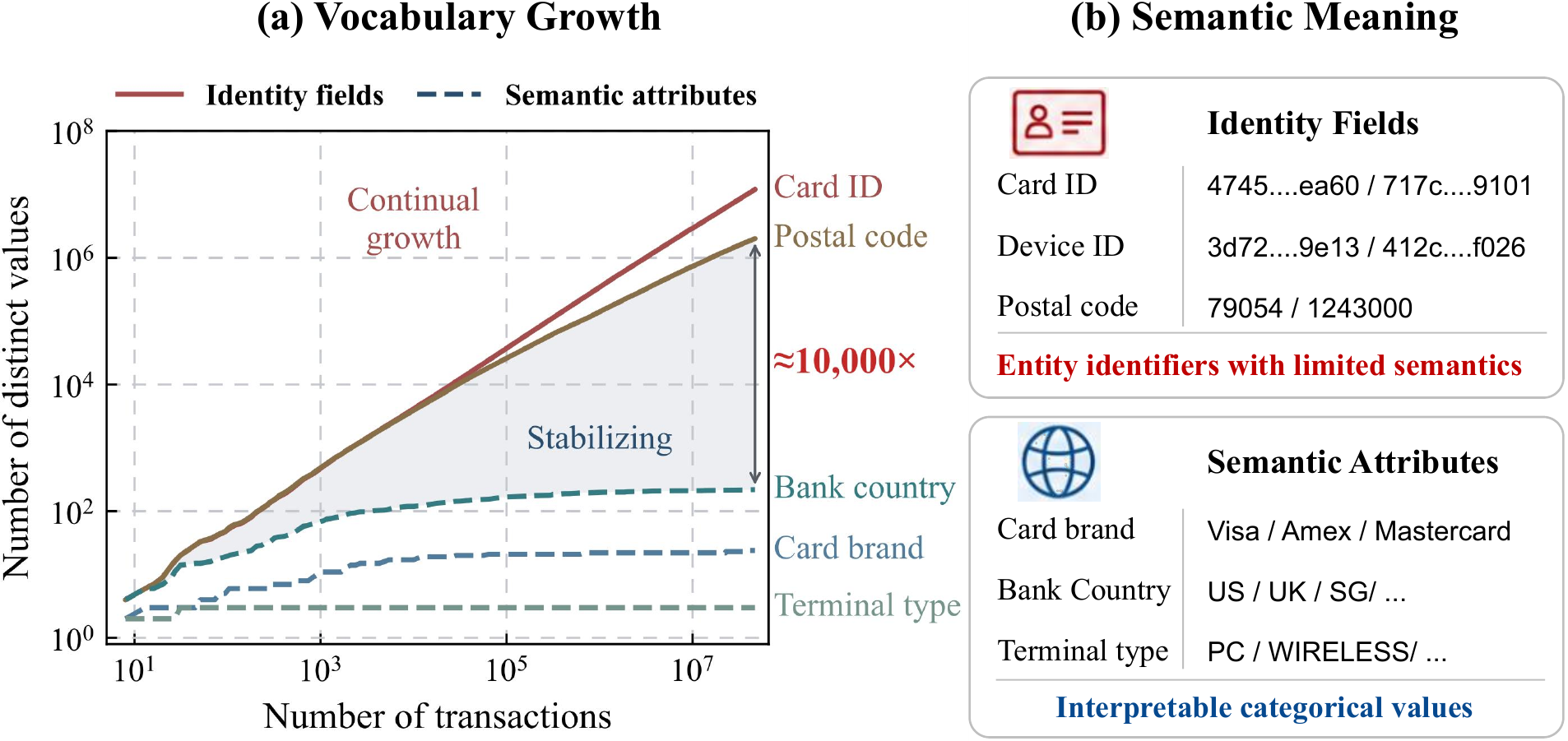}
    \caption{Identity fields vs. semantic attributes on our industrial payment data: (a) continual vocabulary growth; (b) limited semantic meaning.}
    \vspace{-6mm}
    \label{fig:intro}
\end{figure}

Beyond semantic attributes, \textbf{identity fields} such as card, device, and account IDs also carry rich behavioral signals, yet remain underexploited.
These fields characterize users' transaction habits, such as which cards and devices they regularly use, and are therefore critical for risk management in payment systems.
However, as illustrated in Figure~\ref{fig:intro}, identity fields exhibit two characteristics that distinguish them from semantic attributes.
First, \textbf{continual vocabulary growth}: identity fields are high-cardinality, with long-tailed vocabularies that keep expanding as transactions accumulate, whereas semantic vocabularies stabilize.
Second, \textbf{limited semantic meaning}: identity values are arbitrary identifiers rather than interpretable categories like payment types or regions.

These characteristics make existing encoding methods in behavioral sequence models ill-suited to identity fields.
Such methods fall into two categories: \textit{vocabulary-based} and \textit{semantics-based}.
Vocabulary-based methods learn an embedding for each identity value~\cite{sun2019bert4rec,yuan2023where}.
They rely on a fixed vocabulary and generalize poorly to unseen identities, while hashing variants introduce collisions~\cite{kang2021learning, svenstrup2017hash}.
Semantics-based methods encode entity content with pretrained text models~\cite{hou2023learning,rajput2023recommender}.
They avoid vocabulary growth, but identity fields carry little content to encode.
Consequently, neither category effectively represents identity fields.

For high-cardinality identity fields, the behavior signal lies not in specific identity values, but in the \textbf{identity behavior patterns} they exhibit across a transaction sequence.
For example, when a user who has long paid with the same card and device suddenly switches to a new device, the change may indicate account takeover, regardless of the specific device ID.
Such patterns are precisely the behavioral signals that payment risk systems seek from identity fields.
Capturing these patterns from continually growing and semantically opaque identity fields is therefore essential for transaction foundation models to generalize across scenarios and downstream tasks.
However, learning these patterns raises two key challenges:
\begin{itemize}
    \item \textbf{Value-independent identity representation.} How can each identity be represented by its relative identity within the sequence, rather than by its specific value?
    \item \textbf{Sequence-level pattern abstraction.} How can event-level relative identities be abstracted into discrete behavior patterns across multiple temporal ranges?
\end{itemize}

To address these challenges, we propose \method, a pretraining framework that brings identity behavior patterns into transaction foundation models.
Specifically, \method encodes identity fields as relative identities and pretrains on large-scale unlabeled transaction sequences to abstract them into discrete behavior patterns.
For the first challenge, we propose the Local Anonymous ID (LAID) tokenizer, which represents each identity by how it appears and recurs within the sequence, yielding an encoding invariant to specific identity values.
For the second challenge, we design a multi-scale pattern codebook, which abstracts LAID sequences over different temporal ranges into discrete behavior patterns. Experiments on our 100M-scale industrial payment-risk dataset and public financial and commerce datasets show that \method improves 4.13\% on the industrial data and 8.83\% on average across the public datasets.
Notably, \method performs consistently even when all identity values are entirely unseen, demonstrating its potential to generalize to new users and markets.
In summary, the main contributions of our work are as follows:
\begin{itemize}
    \item We identify \textbf{identity behavior patterns}, rather than specific identity values, as the generalizable signal in identity fields, and propose \method to bring them into transaction foundation models.
    \item We propose the LAID tokenizer, which encodes identity fields as relative identities, and a multi-scale pattern codebook, which abstracts them into discrete behavior patterns through self-supervised pretraining.
    \item Experiments on our industrial dataset with over 100M sequences and six public datasets show consistent improvements and stable performance on unseen identity values.
\end{itemize}

\section{Related Work}

\subsection{Transaction Risk Modeling}

Transaction risk models have moved from hand-crafted features toward sequential and graph-based learning.
Early systems feed manually designed statistics, such as velocity features, into tabular learners such as gradient-boosted trees~\cite{chen2016xgboost,prokhorenkova2018catboost}.
Later work models transaction histories directly with sequential networks~\cite{jurgovsky2018sequence,zhang2023fata}, or connects accounts, devices, and merchants into graphs and applies graph neural networks~\cite{liu2018heterogeneous,wang2019semi,dou2020caregnn,liu2021pcgnn,tang2022rethinking}.
Graph methods are where identity fields are used most directly, since shared cards and devices define the edges.
The cost is a global entity graph that must be built and kept up to date as new entities appear.
Our method avoids this by reading identity patterns from each user's own sequence.

\subsection{Pretraining Behavioral Foundation Models}

Pretraining on large-scale behavior sequences has become a common route to foundation models outside language.
In recommendation, BERT4Rec~\cite{sun2019bert4rec} and HSTU~\cite{zhai2024actions} train sequence models with masked and generative objectives, while industrial systems use pretrained knowledge for CTR prediction~\cite{gao2024ppm}, user targeting~\cite{dou2025found}, and generative retrieval~\cite{he2026plum}.
Financial applications follow the same trend, from tabular and contrastive pretraining on transaction sequences~\cite{padhi2021tabular,babaev2022coles} and pretrained models over blockchain transactions~\cite{hu2023bert4eth} to transaction foundation models built for payment fraud detection~\cite{li2025panther,zhang2026friskgpt,dou2025transactiongpt,ostroukhov2026pragma,wu2026transactionfm}.
In these models, however, card and device IDs are typically encoded like any other categorical attribute.
Their continually growing vocabularies and limited semantic meaning receive no special treatment, and this is the gap we address.

\subsection{High-Cardinality ID Representation}

How to represent high-cardinality IDs has been studied mostly in recommendation, where item and user IDs are as arbitrary as transaction identities~\cite{yuan2023where}.
The standard approach learns one embedding per ID~\cite{sun2019bert4rec,zhai2024actions}, and hashing~\cite{weinberger2009feature,svenstrup2017hash}, compositional embeddings~\cite{shi2020compositional}, deep hash embeddings~\cite{kang2021learning}, and frequency-aware compression~\cite{hu2024zipzap} cut the memory cost.
These variants still tie each representation to a specific ID value.
Semantic IDs~\cite{hou2023learning,rajput2023recommender,singh2024better} take another route and quantize item content into shared codes, but this works only because items have titles, images, and descriptions; a card number has none.
The idea closest to ours comes from graph learning, where anonymous walks and related encodings~\cite{ivanov2018anonymous,wang2021inductive,li2020distance} describe nodes by their position in a local structure rather than by their identities.
We carry this relative view over to identity fields in transaction sequences and go one step further, abstracting relative identities into discrete behavior patterns through multi-scale pretraining.

\section{Preliminaries}

\para{Transaction Events and Identity Fields}
Each transaction event consists of identity fields, semantic attributes, and numerical features.
We write the $i$-th event as
$\mathbf{x}_i=(\mathbf{x}_i^{\mathrm{id}},\mathbf{x}_i^{\mathrm{sem}},\mathbf{x}_i^{\mathrm{num}})$,
where $\mathbf{x}_i^{\mathrm{id}}=\{x_{i,f}\}_{f\in\mathcal{F}_{\mathrm{id}}}$
collects the values of the identity fields $\mathcal{F}_{\mathrm{id}}$,
such as card, device, and account IDs.
Each value $x_{i,f}\in\mathcal{V}_f$ identifies the entity associated with field $f$ in the event.
The vocabulary $\mathcal{V}_f$ grows as new entities emerge, and its values are arbitrary identifiers with little intrinsic meaning.

\para{Identity Encoding Paradigms}
Approaches to encoding an identity value $x_{i,f}$ fall into three paradigms, which differ in what the embedding depends on.
\textit{Vocabulary-based encoding} depends on the value itself:
\begin{equation}
    \mathbf{e}_{i,f}^{\mathrm{voc}} = \mathbf{E}_f\big[h_f(x_{i,f})\big],
\end{equation}
where $h_f$ maps an ID to one of $B_f$ rows of the embedding table $\mathbf{E}_f$.
A one-to-one vocabulary lets $B_f$ grow with $|\mathcal{V}_f|$ and maps unseen IDs to a shared out-of-vocabulary row, while hashing fixes $B_f$ at the cost of collisions; in both cases, what the model learns about one ID does not transfer to another ID that behaves the same way.
\textit{Semantics-based encoding} depends on the content associated with the ID:
\begin{equation}
    \mathbf{e}_{i,f}^{\mathrm{sem}} = g_{\phi}\big(c_f(x_{i,f})\big),
\end{equation}
where $c_f(\cdot)$ returns its textual description or attributes and $g_{\phi}$ is a semantic encoder.
Opaque identifiers such as card and device IDs, however, carry little such content.
\textit{Behavior-based encoding} depends only on how the identity relates to others in the sequence:
\begin{equation}
    \mathbf{e}_{i,f}^{\mathrm{beh}} = \psi_f\big(\mathbf{M}_{i,f}\big),
    \quad
    \mathbf{M}_{i,f} = \big[\mathbb{1}[x_{j,f}=x_{k,f}]\big]_{j,k=1}^{i},
    \label{eq:beh}
\end{equation}
where $\mathbf{M}_{i,f}$ records which events up to $i$ share the same identity.
Since $\mathbf{M}_{i,f}$ is unchanged under any one-to-one relabeling of identity values, such encodings apply equally to unseen IDs.
Hand-crafted statistics such as velocity features are predefined choices of $\psi_f$, whereas \method learns $\psi_f$ through LAID and multi-scale pretraining.
Appendix~\ref{app:encoding} compares these paradigms in detail.

\para{Problem Formulation}
For each target transaction, we construct a chronological event sequence
$\mathcal{X}=(\mathbf{x}_1,\ldots,\mathbf{x}_T)$,
where $\mathbf{x}_T$ is the target and $\mathbf{x}_1,\ldots,\mathbf{x}_{T-1}$ is its history.
The task is to predict its risk label $y_T\in\{0,1\}$ using only information available up to time $T$:
\begin{equation}
    \hat{y}_T = F_\theta(\mathcal{X}).
\end{equation}

\begin{figure*}[h]
  \centering
 \includegraphics[width=\linewidth]{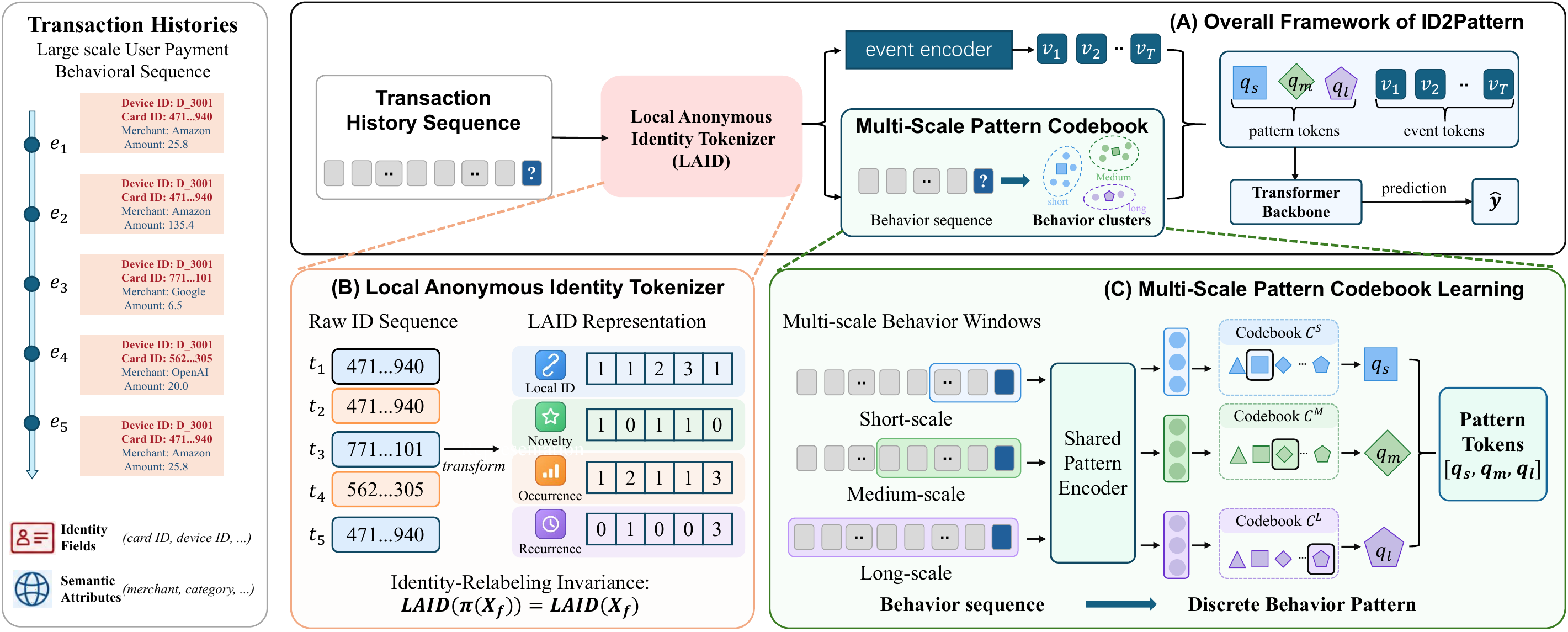}
\caption{Overview of \method. (A) Pattern tokens and event tokens are jointly modeled for risk prediction. (B) The LAID tokenizer encodes each identity by its relative identity within the sequence. (C) The multi-scale pattern codebook quantizes LAID sequences over short, medium, and long windows into discrete pattern tokens.}
\label{fig:framework}
\end{figure*}

\section{Methodology}

\subsection{Overview}
\label{sec:method-overview}
To capture identity behavior patterns from high-cardinality identity fields, we propose \method, which converts raw identity values into discrete behavior patterns and models them together with transaction events.
As illustrated in Figure~\ref{fig:framework}, \method consists of two modules.
The LAID tokenizer encodes each identity by its relative identity within the sequence, yielding a fixed vocabulary that is invariant to specific identity values. The multi-scale pattern codebook then abstracts LAID sequences over different windows into discrete pattern tokens, and is pretrained on unlabeled sequences by recovering masked states. Finally, the pattern tokens are combined with event tokens for downstream risk prediction.

\subsection{Local Anonymous Identity Tokenizer}

\label{sec:laid}

The LAID tokenizer replaces each identity value with a token describing how the identity appears and recurs among other identities in the sequence.
Each identity field $f\in\mathcal{F}_{\mathrm{id}}$ is processed independently.
For event $i$, let
\begin{equation}
p_{i,f}=\min\{j \mid 1\le j\le i,\; x_{j,f}=x_{i,f}\}
\end{equation}
denote the position where the current identity first appears.
The identity is then assigned a local ID
\begin{equation}
\ell_{i,f}=\big|\{x_{j,f}\mid 1\le j\le p_{i,f}\}\big|,
\end{equation}
namely its order of first appearance in the sequence.

The local ID tells which events share the same identity, but not how the identity behaves over time.
We therefore add three behavior descriptors that record its novelty, occurrence, and recurrence:
\begin{equation}
\begin{aligned}
a_{i,f} &= \mathbb{1}[p_{i,f}=i],\\
n_{i,f} &= \textstyle\sum_{j=1}^{i}\mathbb{1}[x_{j,f}=x_{i,f}],\\
d_{i,f} &=
\begin{cases}
0, & p_{i,f}=i,\\
i-\max\{j \mid 1\le j<i,\; x_{j,f}=x_{i,f}\},
& p_{i,f}<i.
\end{cases}
\end{aligned}
\end{equation}
Here, $a_{i,f}$ indicates whether the identity appears for the first time, $n_{i,f}$ counts its occurrences so far, and $d_{i,f}$ is the number of events since its last occurrence, set to $0$ for a new identity.
The four channels form the LAID token $r_{i,f}=(\ell_{i,f},a_{i,f},n_{i,f},d_{i,f})$.

\para{Relabeling Invariance}
All four channels are computed from the equality relations $\mathbf{M}_{i,f}$ and event positions alone, so LAID is a behavior-based encoding in the sense of Eq.~\eqref{eq:beh}.
Hence, for any one-to-one relabeling $\pi_f$ of the values of field $f$,
\begin{equation}
\operatorname{LAID}\big(\pi_f(x_{1,f}),\ldots,\pi_f(x_{T,f})\big)=\operatorname{LAID}\big(x_{1,f},\ldots,x_{T,f}\big).
\end{equation}
A sequence whose identity values are all unseen in training thus receives the same LAID tokens as a seen sequence with the same identity behavior.
Since all channels are bounded by the sequence length, the LAID vocabulary stays fixed as new entities emerge.

\subsection{Multi-Scale Pattern Codebook}
\label{sec:pattern_encoder}

The behavior of an identity depends on the temporal range in which it is viewed: a device that is new within the last few transactions may be familiar over the whole history.
We therefore view each sequence at three scales $\mathcal{S}=\{\mathrm{S},\mathrm{M},\mathrm{L}\}$, corresponding to short, medium, and long windows.
For scale $s$ with window length $L_s$, the window keeps the most recent events,
\begin{equation}
X_s=(\mathbf{x}_{b_s},\ldots,\mathbf{x}_T),\qquad b_s=\max(1,\,T-L_s+1),
\end{equation}
and LAID is recomputed within the window, $R_s=\operatorname{LAID}(X_s)$, so the same identity can receive different LAID tokens at different scales.

\para{Behavior sequence encoding}
Within each window, the four channels of every LAID token are embedded separately and summed over channels and identity fields, giving an event-level behavior embedding
\begin{equation}
\mathbf{e}^{(s)}_i=\sum_{f\in\mathcal{F}_{\mathrm{id}}}\ \sum_{c\in\mathcal{R}} E_{f,c}\big(r^{(s)}_{i,f,c}\big),
\end{equation}
where $\mathcal{R}=\{\mathrm{lid},\mathrm{nov},\mathrm{occ},\mathrm{rcr}\}$ indexes the four channels and $E_{f,c}$ is a channel-specific embedding table.
A Transformer encoder~\cite{vaswani2017attention} shared across scales encodes the behavior sequence $(\mathbf{e}^{(s)}_{b_s},\ldots,\mathbf{e}^{(s)}_T)$ of each window, and its output at the current event, denoted $\mathbf{h}_s$, summarizes the identity behavior within scale $s$.

\para{Pattern abstraction}
The continuous representation $\mathbf{h}_s$ still varies from sequence to sequence.
To abstract it into a finite set of behavior patterns shared across users, each scale maintains its own codebook $\mathcal{C}^{(s)}=\{\mathbf{c}^{(s)}_1,\ldots,\mathbf{c}^{(s)}_K\}$, and $\mathbf{h}_s$ is replaced by its nearest code:
\begin{equation}
\mathbf{q}_s=\mathbf{c}^{(s)}_{k^*},\qquad k^*=\arg\min_{k}\big\|\mathbf{h}_s-\mathbf{c}^{(s)}_k\big\|_2 .
\label{eq:code}
\end{equation}
$\mathbf{q}_s$ denotes the pattern token of scale $s$.
The three tokens $[\mathbf{q}_{\mathrm{S}},\mathbf{q}_{\mathrm{M}},\mathbf{q}_{\mathrm{L}}]$ summarize identity behavior from complementary temporal ranges.

\subsection{Self-Supervised Pattern Pretraining}
\label{sec:pretraining}
To learn reusable behavior patterns without labels, we pretrain the pattern branch, i.e., the behavior embeddings, pattern encoder, and codebooks, on training-period sequences by recovering the hidden identity states of the current transaction.
This exploits the correlation among identities within a transaction: whether a device is new is closely related to whether the accompanying card is familiar.

\para{Masked identity-state recovery}
For each sequence, we mask the LAID tokens of a random nonempty field subset $\mathcal{F}_{\mathrm{mask}}\subseteq\mathcal{F}_{\mathrm{id}}$ at $\mathbf{x}_T$ in all three windows.
With the whole token masked, including its local ID, the model must infer these states from how the other identities of $\mathbf{x}_T$ relate to the history.
For each scale $s$ and field $f\in\mathcal{F}_{\mathrm{mask}}$, a field-specific head predicts the novelty $a^{(s)}_{T_s,f}$, occurrence $n^{(s)}_{T_s,f}$, and recurrence $d^{(s)}_{T_s,f}$ from $\mathbf{q}_s$, where $T_s$ is the position of $\mathbf{x}_T$ in $X_s$, so each code must capture the behavior needed for recovery.

\para{Pretraining objective}
Novelty is binary and optimized with binary cross-entropy, while occurrence and recurrence are heavy-tailed and regressed in log space, $g(x)=\log(1+x)$:
\begin{equation}
\begin{aligned}
\mathcal{L}^{(s,f)}_{\mathrm{nov}}&=\operatorname{BCE}\big(\hat{a}^{(s)}_{T_s,f},\,a^{(s)}_{T_s,f}\big),\\
\mathcal{L}^{(s,f)}_{\mathrm{occ}}&=\operatorname{SmoothL1}\big(\hat{n}^{(s)}_{T_s,f},\,g(n^{(s)}_{T_s,f})\big),\\
\mathcal{L}^{(s,f)}_{\mathrm{rcr}}&=\operatorname{SmoothL1}\big(\hat{d}^{(s)}_{T_s,f},\,g(d^{(s)}_{T_s,f})\big),
\end{aligned}
\end{equation}
where $\hat{n}$ and $\hat{d}$ denote predictions in log space.
The recovery loss averages over masked fields and scales:
\begin{equation}
\label{eq:recover}
\mathcal{L}_{\mathrm{recover}}=\frac{1}{|\mathcal{S}||\mathcal{F}_{\mathrm{mask}}|}\sum_{s\in\mathcal{S}}\sum_{f\in\mathcal{F}_{\mathrm{mask}}}\Big(\lambda_a\mathcal{L}^{(s,f)}_{\mathrm{nov}}+\lambda_n\mathcal{L}^{(s,f)}_{\mathrm{occ}}+\lambda_d\mathcal{L}^{(s,f)}_{\mathrm{rcr}}\Big).
\end{equation}
The codebooks are trained with the VQ-VAE objective~\cite{oord2017vqvae}, using straight-through gradients for the nearest-code selection:
\begin{equation}
\mathcal{L}_{\mathrm{VQ}}=\sum_{s\in\mathcal{S}}\Big(\big\|\operatorname{sg}[\mathbf{h}_s]-\mathbf{q}_s\big\|_2^2+\beta\big\|\mathbf{h}_s-\operatorname{sg}[\mathbf{q}_s]\big\|_2^2\Big),
\label{eq:vq}
\end{equation}
where $\operatorname{sg}[\cdot]$ is the stop-gradient operator and $\beta$ weights the commitment term.
The pretraining loss is $\mathcal{L}_{\mathrm{pre}}=\mathcal{L}_{\mathrm{recover}}+\lambda_{\mathrm{VQ}}\mathcal{L}_{\mathrm{VQ}}$, plus a code-balancing regularizer against codebook collapse (Appendix~\ref{app:impl}).
The recovery head is then discarded, and the pattern branch is transferred to downstream modeling.

\subsection{Pattern-Augmented Downstream Modeling}
\label{sec:downstream}

To exploit the pretrained patterns in risk prediction, we combine the pattern tokens with event tokens that retain transaction-level details.

\para{Event representation}
\method encodes every field of an event through a finite vocabulary.
Each semantic attribute $f \in \mathcal{F}_{\mathrm{sem}}$ has a stable vocabulary and is embedded by its own table, $\mathbf{e}^{\mathrm{sem}}_{i,f} = \mathbf{E}^{\mathrm{sem}}_{f}(x_{i,f})$.
Each identity value is instead replaced by its LAID token $r_{i,f}$ computed over the full sequence $\mathcal{X}$, whose bounded channels form fixed vocabularies:
\begin{equation}
\mathbf{e}^{\mathrm{id}}_{i,f} = \sum_{c \in \mathcal{R}} \mathbf{E}^{\mathrm{id}}_{f,c}\big(r_{i,f,c}\big).
\end{equation}
Numerical features are linearly projected, $\mathbf{e}^{\mathrm{num}}_{i} = \mathbf{W}_{\mathrm{num}} \mathbf{x}^{\mathrm{num}}_{i}$, and the event token concatenates and projects all field embeddings:
\begin{equation}
\mathbf{v}_i = \mathbf{W}_{v} \big[\, \{\mathbf{e}^{\mathrm{id}}_{i,f}\}_{f \in \mathcal{F}_{\mathrm{id}}};\ \{\mathbf{e}^{\mathrm{sem}}_{i,f}\}_{f \in \mathcal{F}_{\mathrm{sem}}};\ \mathbf{e}^{\mathrm{num}}_{i} \,\big].
\label{eq:event}
\end{equation}
This event branch keeps transaction-level details, such as amounts and categories, that the pattern tokens do not capture.

\para{Pattern-augmented prediction}
The pattern tokens are prepended to the event sequence and modeled jointly by a Transformer backbone:
\begin{equation}
\begin{aligned}
\mathbf{H}&=\operatorname{Backbone}\big([\mathbf{q}_{\mathrm{S}},\mathbf{q}_{\mathrm{M}},\mathbf{q}_{\mathrm{L}},\mathbf{v}_1,\ldots,\mathbf{v}_T]\big),\\
\hat{y}_T&=\operatorname{Head}\big(\operatorname{Readout}(\mathbf{H})\big),
\end{aligned}
\end{equation}
where $\operatorname{Readout}$ takes the output at the target event $\mathbf{x}_T$.
The model is trained with binary cross-entropy between $\hat{y}_T$ and $y_T$.
The event encoder, backbone, and head are randomly initialized, while the pattern branch is initialized from pretraining and fine-tuned jointly; the LAID tokenizer is deterministic and shared by both stages.

\subsection{Complexity}
\method adds a constant-factor overhead to the backbone and is independent of the number of identities.
The LAID tokenizer computes the four channels in a single pass over each window with a hash map, taking $O(|\mathcal{F}_{\mathrm{id}}| \sum_{s} L_s)$ time; Algorithm~\ref{alg:id2pattern} in Appendix~\ref{app:impl} summarizes the full procedure.
The shared pattern encoder costs $O(\sum_{s} L_s^2 d)$, dominated by the long window, and the codebook lookup costs $O(|\mathcal{S}| K d)$.
The backbone processes only three additional tokens, so its cost remains $O(T^2 d)$.
None of these terms depends on the vocabulary size $|\mathcal{V}_f|$, which keeps both the computation and the parameters of identity representation fixed as new entities emerge.

\section{Experiments}

We evaluate the effectiveness of \method through extensive experiments, guided by the following research questions:
\begin{itemize}
    \item \textbf{RQ1:} How effective is \method compared with strong baselines on industrial and public benchmarks?
    \item \textbf{RQ2:} How does each component of \method contribute to its overall effectiveness?
    \item \textbf{RQ3:} How robust is \method to entirely unseen identities, and what behavior patterns do the codebooks learn?
    \item \textbf{RQ4:} Do the learned identity behavior patterns generalize across different backbones and transaction datasets?
\end{itemize}

\subsection{Experimental Setup}

\begin{table}[t]
\centering
\caption{Statistics of evaluated datasets. \#Sequences counts the supervised targets, each with its own history sequence; the positive rate is computed on the test set.}
\label{tab:datasets}
\small
\begin{tabular}{@{}lcrcr@{}}
\toprule
Dataset & Domain & \#Sequences & \#ID Fields & Pos. Rate \\
\midrule
Industrial   & Risk     & 100.0M+ & 9 & --      \\
Elliptic++   & Risk     & 37.3K   & 2 & 5.69\%  \\
IBM AML      & Risk     & 1.54M   & 2 & 0.20\%  \\
CCT          & Risk     & 7.44M   & 2 & 0.12\%  \\
RetailRocket & Commerce & 540.4K  & 1 & 2.78\%  \\
Synerise     & Commerce & 6.12M   & 2 & 24.44\% \\
Taobao       & Commerce & 39.26M  & 2 & 1.88\%  \\
\bottomrule
\end{tabular}
\end{table}

\para{Datasets}
We evaluate \method on one industrial payment-risk dataset and six public datasets covering financial risk and commerce behavior.
The industrial dataset contains over 100M transaction sequences from a real-world payment platform with nine identity fields, such as card, device, and account IDs, and is split chronologically into training, validation, and test periods.
The public financial-risk datasets are CCT~\cite{padhi2021tabular}, IBM AML~\cite{altman2023amlworld}, and Elliptic++~\cite{elmougy2023ellipticpp}, which label each transaction as fraudulent, laundering, or illicit.
The commerce datasets are Taobao~\cite{tianchiTaobao}, RetailRocket~\cite{retailrocketDataset}, and Synerise~\cite{synerise2025dataset}, which test whether identity behavior patterns remain useful beyond risk detection: Taobao and RetailRocket predict whether the current event is a purchase, and Synerise predicts whether an add-to-cart event converts into a purchase.
Table~\ref{tab:datasets} summarizes the statistics, and Appendix~\ref{app:datasets} details the preprocessing.

\para{Baselines}
\textit{Industrial evaluation.}
We use the existing internal production sequence foundation model as the baseline, which encodes semantic attributes and numerical features but no identity fields, denoted as \textit{Semantic-attribute-only} in Table~\ref{tab:industrial}.
To isolate the effect of identity representation, we add identity fields to this model through hashed embeddings (\textit{+ Hash ID}) or LAID (\textit{+ LAID}), and \method further adds the pretrained pattern tokens.
The \textit{relabeled} variants evaluate the same trained models after replacing every test-set identity value with an unseen one, while preserving which events share the same identity.

\textit{Public benchmarks.}
We compare \method with three groups of baselines: (1) sequential models, including LSTM, GRU, and Transformer; (2) tabular and boosting models, including XGBoost~\cite{chen2016xgboost}, CatBoost~\cite{prokhorenkova2018catboost}, and Temporal Boosting~\cite{gupta2025temporalboosting}; and (3) pretraining-based and specialized models, including BERT4Rec~\cite{sun2019bert4rec}, TabBERT~\cite{padhi2021tabular}, HSTU\cite{zhai2024actions}, PANTHER~\cite{li2025panther}, IBM AI-on-Z GRU~\cite{ibmAiOnZFraud}, NVIDIA TFM~\cite{wu2026transactionfm}, FRiskGPT~\cite{zhang2026friskgpt}, and PRAGMA~\cite{ostroukhov2026pragma}. All baselines are trained on our datasets, and Appendix~\ref{app:baselines} provides further details.

\begin{table}[t]
\centering
\caption{Relative performance changes on the industrial payment-risk
test set, using the semantic-attribute-only baseline as the reference.}
\label{tab:industrial}
\small
\begin{tabular}{lccc}
\toprule
Method & AUC-ROC (\%) & T1 (\%) & AddCB (\%) \\
\midrule
Semantic-attribute-only & --- & --- & --- \\
 + Hash ID & $-5.87$ & $-11.91$ & $-8.07$ \\
 + Hash ID (relabeled) & $-8.00$ & $-50.53$ & $-46.96$ \\
 + LAID & $+1.11$ & $+6.96$ & $+6.09$ \\
 + LAID (relabeled) & $+1.11$ & $+6.96$ & $+6.09$ \\
\midrule
\rowcolor{blue!5}
\textbf{\method} & \textbf{+4.13} & \textbf{+22.64} & \textbf{+19.67} \\
\bottomrule
\end{tabular}
\end{table}
\begin{figure}[ht]
  \centering
  \includegraphics[width=\linewidth]{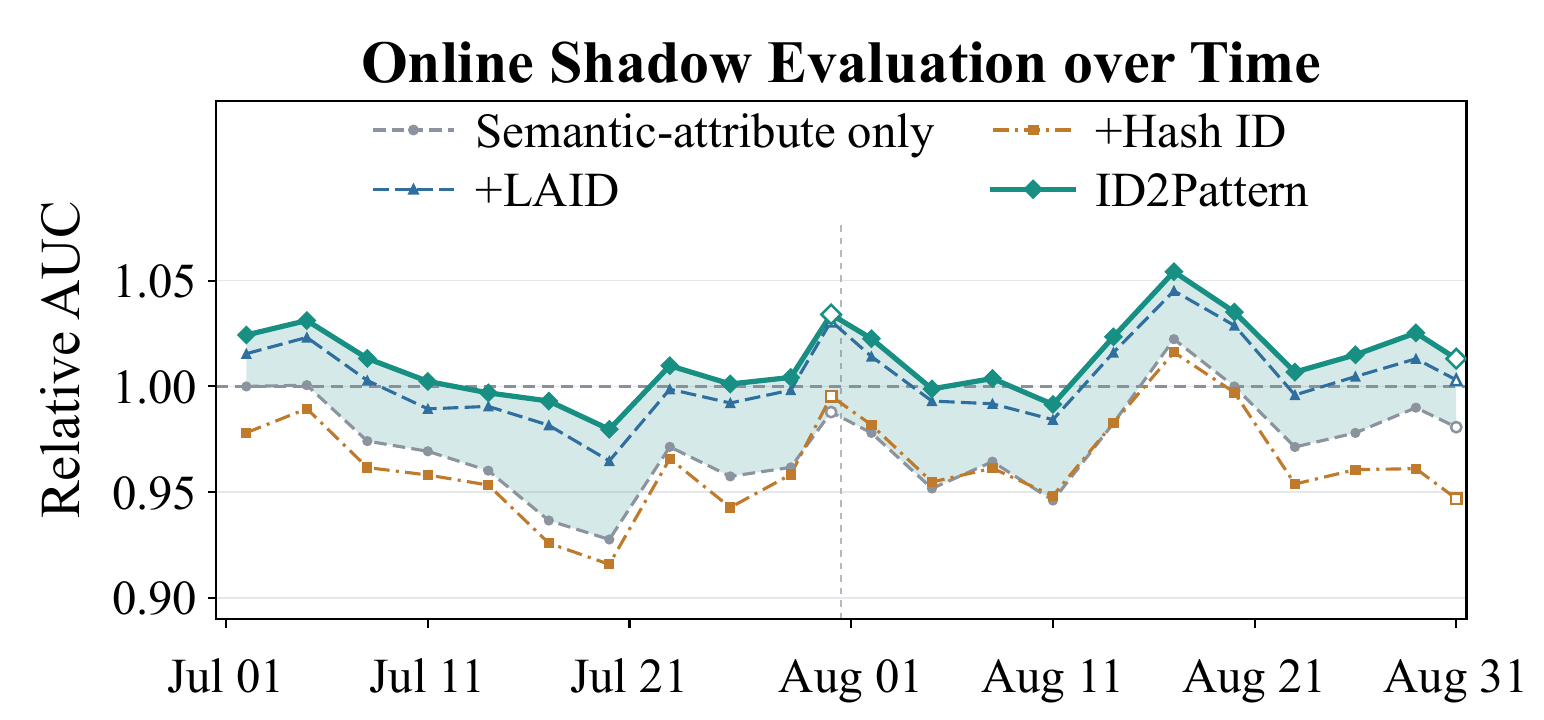}
    \caption{Online shadow evaluation from July to August. AUC is normalized by the average AUC of the semantic-attribute-only model over the first three days.}
  \label{fig:online}
\end{figure}

\begin{table*}[t]
\centering
\caption{Performance on six public datasets in terms of PR-AUC (PR) and recall at a false positive rate of at most 0.5\% (R@0.5\%). Best results are in bold, and second-best results are underlined.}
\label{tab:public}
\small
\setlength{\tabcolsep}{2.0pt}
\renewcommand{\arraystretch}{1.00}
\resizebox{0.95\textwidth}{!}{
\begin{tabular}{lcccccccccccc}
\toprule
& \multicolumn{6}{c}{\textbf{Risk}}
& \multicolumn{6}{c}{\textbf{Commerce}} \\
\cmidrule(lr){2-7}
\cmidrule(lr){8-13}

& \multicolumn{2}{c}{Elliptic++}
& \multicolumn{2}{c}{IBM AML}
& \multicolumn{2}{c}{CCT}
& \multicolumn{2}{c}{RetailRocket}
& \multicolumn{2}{c}{Synerise}
& \multicolumn{2}{c}{Taobao} \\
\cmidrule(lr){2-3}
\cmidrule(lr){4-5}
\cmidrule(lr){6-7}
\cmidrule(lr){8-9}
\cmidrule(lr){10-11}
\cmidrule(lr){12-13}

Method
& PR & R@0.5\%
& PR & R@0.5\%
& PR & R@0.5\%
& PR & R@0.5\%
& PR & R@0.5\%
& PR & R@0.5\% \\
\midrule

\multicolumn{13}{l}{\textit{Sequential Models}} \\

LSTM
& .2410 & .0697
& .2014 & .3436
& .0318 & .0665
& .7255 & .5168
& .4411 & \underline{.0341}
& .2078 & .1528 \\

GRU
& .2428 & .0860
& .1600 & .3594
& .0315 & .0675
& .7358 & .5685
& \underline{.4487} & .0337
& .2102 & .1553 \\

Transformer
& .2333 & .0540
& .2129 & .3800
& .0303 & .0873
& .7518 & .5695
& .4138 & .0214
& .2026 & .1480 \\

\midrule
\multicolumn{13}{l}{\textit{Tabular / Boosting Models}} \\

XGBoost
& .1691 & .0409
& .1269 & \underline{.4393}
& .0097 & .0287
& .6398 & .4841
& .4382 & .0276
& \underline{.2368} & \underline{.1656} \\

CatBoost
& .1851 & .0456
& .1481 & .3973
& .0158 & .0439
& .6476 & .5096
& .4373 & .0273
& .2012 & .1405 \\

Temporal Boosting
& .1614 & .0566
& .0826 & .0233
& .0130 & .0164
& .5962 & .4636
& .4251 & .0242
& .2078 & .1507 \\

\midrule
\multicolumn{13}{l}{\textit{Pretrained / Foundation Models}} \\

BERT4Rec
& .2414 & .0781
& .2398 & .4280
& \underline{.0562} & .2040
& .7651 & \underline{.6239}
& .4142 & .0230
& .2029 & .1482 \\

TabBERT
& .2125 & .0613
& .1316 & .2264
& .0375 & .1513
& .7604 & .6088
& .3950 & .0194
& .2260 & .1471 \\

HSTU
& \underline{.2489} & \underline{.0917}
& .0955 & .2962
& .0228 & .0581
& .7608 & .6148
& .4176 & .0230
& .2050 & .1514 \\

PANTHER
& .1963 & .0697
& .2543 & .1486
& .0250 & .0569
& .6629 & .5793
& .4471 & .0266
& .1837 & .1178 \\

IBM AI-on-Z GRU
& .2381 & .0786
& \underline{.2588} & .3902
& .0291 & .0595
& .7472 & .5710
& .3551 & .0260
& .1932 & .1441 \\

NVIDIA TFM
& .1622 & .0225
& .1430 & .2711
& .0483 & \underline{.2044}
& .7531 & .5999
& .4198 & .0221
& .1988 & .1443 \\

FRiskGPT
& .1553 & .0142
& .1982 & .3673
& .0487 & .1903
& \underline{.7698} & \underline{.6239}
& .4227 & .0230
& .2024 & .1473 \\
PRAGMA
& .2409 & .0204
& .1774 & .1067
& .0197 & .0339
& .6181 & .4262
& .3476 & .0167
& .2245 & .1570\\
\midrule
\rowcolor{blue!5}

\textbf{\method}
& \textbf{.2758} & \textbf{.1001}
& \textbf{.2761} & \textbf{.4951}
& \textbf{.0607} & \textbf{.2161}
& \textbf{.8283} & \textbf{.6903}
& \textbf{.4776} & \textbf{.0464}
& \textbf{.2686} & \textbf{.1845} \\

\bottomrule
\end{tabular}
}
\end{table*}

\para{Metrics and implementation}
On the public datasets, we report PR-AUC as the primary metric and R@0.5\%, the recall at a false positive rate of at most 0.5\% on the test set.
On the industrial dataset, we report AUC-ROC and two business metrics: T1, the recall among the top 1\% highest-scored transactions, and AddCB, the additional chargeback recall among transactions not intercepted by the existing risk system.
\method uses windows of 7, 30, and $L_{\max}$ events with $K = 64$ codes per scale, and each public experiment is repeated with three random seeds.
Appendix~\ref{app:industrial} details the industrial metrics and online protocol, and Appendix~\ref{app:impl} lists the remaining hyperparameters.

\subsection{Industrial Payment-Risk Evaluation (\textbf{RQ1})}

We first evaluate \method on industrial payment risk through offline experiments on over 100M transaction sequences and a two-month online shadow evaluation on live traffic.

\para{Offline evaluation}
As illustrated in Table~\ref{tab:industrial}, \method achieves the best results on all three metrics, improving AUC-ROC, T1, and AddCB by 4.13\%, 22.64\%, and 19.67\% over the semantic-attribute-only model.
The controlled variants show that these gains depend on how identity fields are represented.
Adding identity fields through hashing degrades all metrics below the reference model, and relabeling test identities with unseen values further reduces its T1 by 43.8\%, revealing its reliance on specific identity values.
In contrast, LAID improves over the reference model and remains unchanged under relabeling, and the pattern tokens bring further gains on top of LAID.

\para{Online shadow evaluation}
We further evaluate \method in shadow mode on live production traffic from July to August. Since risk labels such as chargebacks arrive weeks after transactions, we compute the metrics after the labels mature rather than measuring real-time business gains (Appendix~\ref{app:industrial}).
As shown in Figure~\ref{fig:online}, the offline gains persist on live traffic: \method achieves the highest AUC on every day, and the variants keep the same order as offline, with LAID above hashing and the pattern tokens adding a consistent margin over LAID.

\subsection{Public Benchmark Evaluation (\textbf{RQ1})}
To evaluate whether the gains of \method generalize beyond our platform, we compare it with 14 baselines on three financial-risk and three commerce datasets.
As shown in Table~\ref{tab:public}, we make three observations.

\para{Overall performance}
First, \method consistently outperforms all baselines, while no baseline is consistently strong.
We achieve the best PR-AUC and R@0.5\% on all six datasets, improving PR-AUC by 8.83\% on average over the strongest baseline on each dataset.
The strongest baseline differs on every dataset, and generic pretraining alone is insufficient: HSTU performs well on Elliptic++ but falls below GRU and CatBoost on IBM AML.
These results suggest that the gains of \method come from how identity fields are represented rather than from pretraining or model capacity.

\para{Financial-risk tasks}
Second, \method is effective at ranking the riskiest transactions.
On the three risk datasets, it improves PR-AUC by 6.7--10.8\% and R@0.5\% by 5.7--12.7\% over the strongest baseline, where R@0.5\% reflects how many risky transactions are caught while at most 0.5\% of legitimate ones are flagged.
Tabular models, which summarize the history into aggregated statistics, fall far behind on CCT, suggesting that the order in which identities are reused or switched carries risk signals that aggregation discards.

\para{Commerce tasks}
Third, identity behavior patterns remain useful beyond risk detection.
On the three commerce datasets, \method improves PR-AUC by 6.4--13.4\% over the strongest baseline, with the largest gain on Taobao.
Purchase intent is often revealed by repeated visits to the same item within a short period, as shown by the revisit pattern on Taobao in Section~\ref{sec:rq3}, which the short-window codes of \method capture directly.

\subsection{Ablation Study (\textbf{RQ2})}
To understand how each component contributes, we compare our method with six variants on two risk and two commerce datasets.
\textbf{w/o LAID} replaces LAID with hash-based identity encoding;
\textbf{Local ID only} keeps only the local-ID channel of LAID;
\textbf{w/o Pattern Branch} removes the pattern tokens and keeps only the event branch;
\textbf{Single-scale} uses only the long window;
\textbf{w/o Codebook} feeds continuous representations $\mathbf{h}_s$ instead of discrete pattern tokens;
and \textbf{w/o Pretraining} trains the full model from scratch.

As shown in Table~\ref{tab:ablation}, every component contributes, and LAID contributes the most.
Removing any component lowers PR-AUC on all four datasets, and replacing LAID with hashing causes the largest average drop, reaching 44.3\% on CCT.
Even within LAID, the novelty, occurrence, and recurrence channels improve over local IDs alone on all four datasets.
The pattern branch, multi-scale windows, codebook, and pretraining each bring further gains on top of LAID.
Notably, discrete pattern tokens outperform their continuous counterparts on all four datasets, which supports abstracting identity behavior into a finite set of shared patterns rather than keeping sequence-specific representations.

The pattern branch matters more for risk tasks than for commerce tasks.
On IBM AML, removing the pattern branch, multi-scale windows, codebook, or pretraining reduces PR-AUC by 15\% to 32\%, whereas on RetailRocket and Synerise the drops stay within 5\%.
In contrast, replacing LAID with hashing still costs 5.7\% and 11.5\% on these two commerce datasets.
This contrast is consistent with our motivation: how cards, devices, and accounts are reused or switched carries strong risk signals that call for multi-scale pattern abstraction, while in commerce the LAID-encoded event branch already captures most of the purchase signal.
\begin{table}[H]
\centering
\caption{Ablation study of different components (PR-AUC).}
\label{tab:ablation}
\small
\setlength{\tabcolsep}{4pt}
\renewcommand{\arraystretch}{1.05}
\resizebox{\columnwidth}{!}{
\begin{tabular}{lcccc}
\toprule
\multirow{2}{*}{\textbf{Ablation Variants}}
& \multicolumn{2}{c}{\textbf{Risk}}
& \multicolumn{2}{c}{\textbf{Commerce}} \\
\cmidrule(lr){2-3}
\cmidrule(lr){4-5}
& CCT & IBM AML & RetailRocket & Synerise \\
\midrule

w/o LAID
& .0338 & .2018 & .7814 & .4228\\

Local ID Only
& .0600 & .2695 & .8152 & .4718 \\

w/o Pattern Branch
& .0418 & .1880 & .8064 & .4734 \\

Single-scale
& .0526 & .2263 & .8087 & .4700 \\

w/o Codebook
& .0419 & .2047 & .8092 & .4630 \\

w/o Pretraining
& .0451 & .2351 & .8011 & .4561 \\

\midrule
\textbf{\method}
& \textbf{.0607}
& \textbf{.2761}
& \textbf{.8283}
& \textbf{.4776} \\

\bottomrule
\end{tabular}
}
\end{table}

\subsection{Identity Generalization and Pattern Analysis (\textbf{RQ3})}
\label{sec:rq3}
To evaluate whether \method generalizes to new identities and what identity behavior patterns it learns, we conduct a \textbf{generalization study} under identity relabeling and a \textbf{case study} on the learned codebooks.

\begin{figure}[H]
  \centering
  \includegraphics[width=\linewidth]{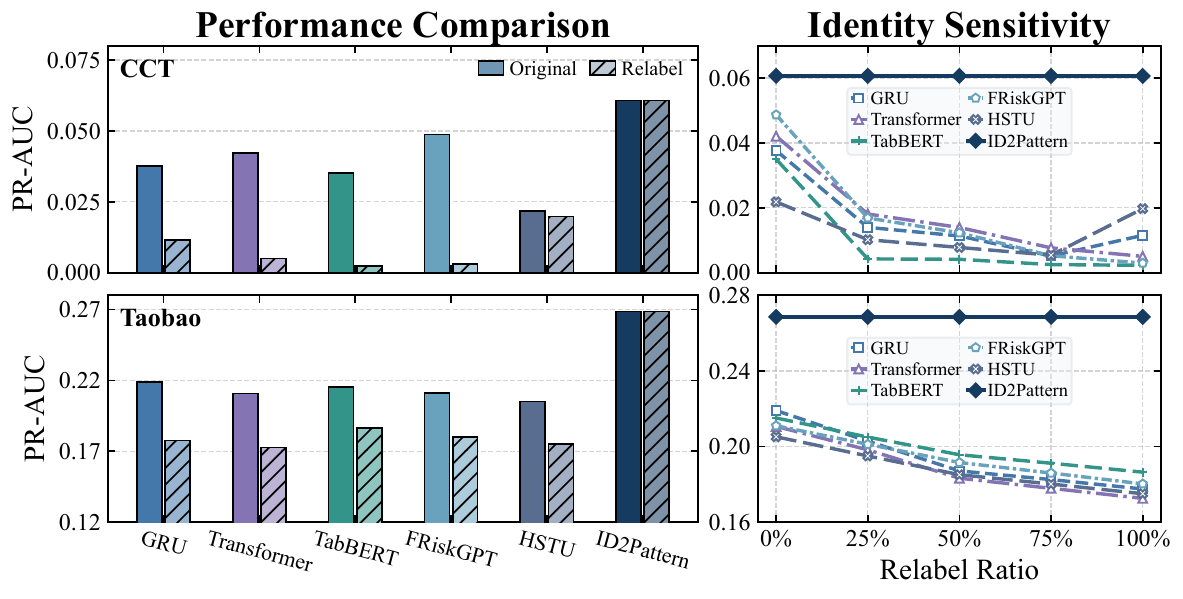}
  \caption{Generalization to unseen identities on CCT and Taobao. Left: PR-AUC with original and fully relabeled identities. Right: PR-AUC under increasing relabeling ratios.}
  \label{fig:identity_robustness}
\end{figure}

\para{Generalization to unseen identities}
To test whether models rely on specific identity values, we relabel identity values in the test set with values never seen during training, while keeping which events share the same identity unchanged.
For this study, we additionally train five representative baselines with hashed identity fields as inputs, keeping all other settings unchanged.
We relabel either all identities or a growing fraction of them, from 0\% to 100\%, and evaluate all models without retraining.
As shown in Figure~\ref{fig:identity_robustness}, \method keeps exactly the same PR-AUC under every ratio, as guaranteed by the relabeling invariance of LAID.
In contrast, the baselines already lose 53.4--87.9\% of their PR-AUC on CCT and 4.7--7.4\% on Taobao at 25\% relabeling; some partly recover at 100\% on CCT, most notably HSTU, as no memorized identity remains to conflict with the unseen ones.
The drop is larger on CCT, where fraud depends more on whether merchants and cities are familiar.

\begin{figure}[t]
  \centering
  \includegraphics[width=0.98\linewidth]{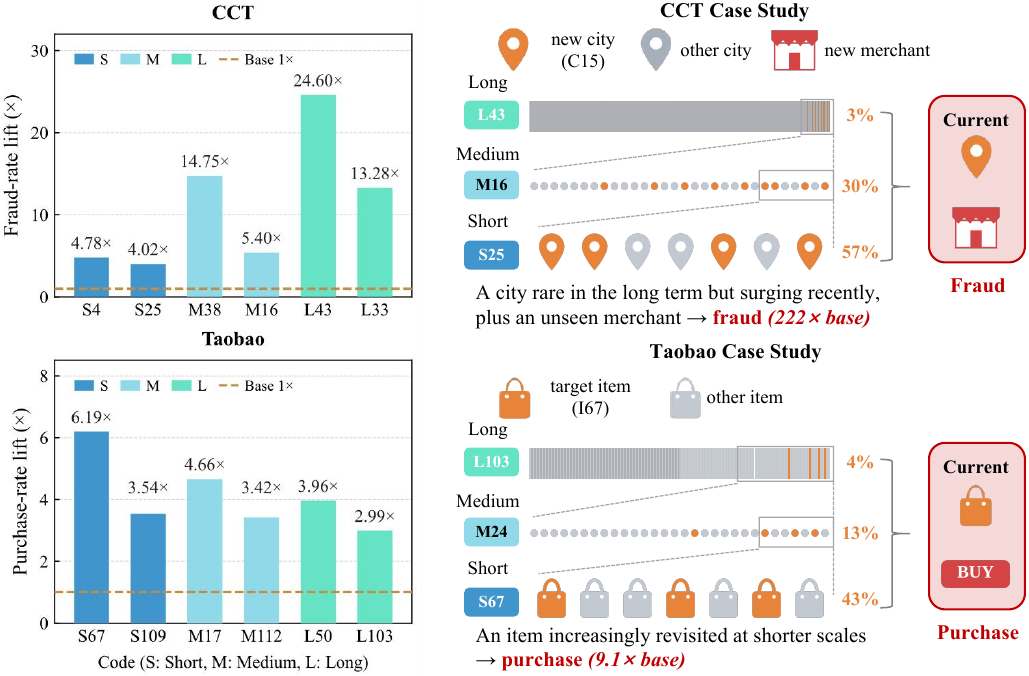}
  \caption{Case study on learned codebooks. Left: positive-rate lift of codes at each scale. Right: multi-scale code combinations describing behavior patterns on CCT and Taobao.}
  \label{fig:pattern_case}
\end{figure}

\para{Case study on learned codebooks}
To examine whether the codes carry behavioral meaning, we group test samples by the code assigned at each scale and compare the positive rate of each group with the dataset average. As shown in Figure~\ref{fig:pattern_case} (left), several codes concentrate positive samples far above the base rate, reaching 24.6$\times$ the fraud rate on CCT and 6.2$\times$ the purchase rate on Taobao.
The most informative scale also differs by task: long-window codes are the strongest on CCT and short-window codes on Taobao, suggesting that fraud deviates from long-term habits while purchases reflect recent intent.

Combining codes across scales further reveals interpretable behavior patterns (Figure~\ref{fig:pattern_case}, right).
On CCT, the codes describe a city rare in the long history but accounting for 30\% of the last 30 events and 57\% of the last 7, followed by a payment to an unseen merchant; this combination has a fraud rate 222$\times$ the base rate.
On Taobao, they describe an item whose share rises from 4\% to 13\% and 43\% as the window shrinks, followed by its purchase, with a purchase rate 9.1$\times$ the base rate.
These patterns capture two distinct identity behaviors, emerging novelty and repeated revisits, and depend only on how identities appear relative to the history, not on which city or item is involved.

\subsection{Representation Generalizability (\textbf{RQ4})}
The identity modeling of \method generalizes beyond the architecture and dataset on which it is developed.
We examine this property from two aspects: generalization across sequence backbones and transfer across datasets.

\para{Generalization across Backbones}
\method improves diverse sequence backbones without modifying their architectures.
To test this, we attach the identity modeling of \method, namely LAID encoding and the pretrained pattern tokens, to four representative backbones, GRU, LSTM, Transformer, and Mamba, and compare each with its original version.
As shown in Table~\ref{tab:backbone}, \method improves PR-AUC in 15 of the 16 backbone--dataset combinations, covering recurrent, attention-based, and state-space architectures.
The gains are consistent on the commerce datasets, ranging from 4.6\% to 11.4\% on RetailRocket and from 26.9\% to 32.6\% on Taobao.
On the risk datasets, the gains are larger but less uniform, ranging from $-15.7\%$ to $+100.3\%$, as PR-AUC fluctuates more at the low positive rates of IBM AML and CCT.

\definecolor{ptbg}{HTML}{EEF0FA}    %
\definecolor{gain}{HTML}{2E7D32}
\definecolor{loss}{HTML}{C62828}
\newcommand{\up}[1]{\textcolor{gain}{+#1\%}}
\newcommand{\down}[1]{\textcolor{loss}{$-$#1\%}}

\begin{table}[t]
\centering
\caption{PR-AUC of different backbones with and without the identity modeling of \method, averaged over 3 seeds.}
\label{tab:backbone}
\small
\setlength{\tabcolsep}{5pt}
\begin{tabular}{@{}lcccc@{}}
\toprule
 & \multicolumn{2}{c}{Risk} & \multicolumn{2}{c}{Commerce} \\
\cmidrule(lr){2-3}\cmidrule(lr){4-5}
Backbone & IBM AML & CCT & RetailRocket & Taobao \\
\midrule
GRU                              & .1600 & .0315 & .7358 & .2102 \\
\rowcolor{ptbg} \quad + \method  & .2397 & .0345 & .7700 & .2756 \\
\quad \textit{Improv.}           & \up{49.8} & \up{9.5} & \up{4.6} & \up{31.1} \\
\midrule
LSTM                             & .2014 & .0318 & .7255 & .2078 \\
\rowcolor{ptbg} \quad + \method  & .2543 & .0268 & .7763 & .2678 \\
\quad \textit{Improv.}           & \up{26.3} & \down{15.7} & \up{7.0} & \up{28.9} \\
\midrule
Transformer                      & .2129 & .0303 & .7518 & .2026 \\
\rowcolor{ptbg} \quad + \method  & .2761 & .0607 & .8283 & .2686 \\
\quad \textit{Improv.}           & \up{29.7} & \up{100.3} & \up{10.2} & \up{32.6} \\
\midrule
Mamba                            & .1999 & .0285 & .7033 & .1991 \\
\rowcolor{ptbg} \quad + \method  & .2133 & .0390 & .7833 & .2526 \\
\quad \textit{Improv.}           & \up{6.7} & \up{36.8} & \up{11.4} & \up{26.9} \\
\bottomrule
\end{tabular}
\end{table}

\para{Generalization across Datasets}
Pattern encoders pretrained on other datasets transfer to a new dataset and remain as effective as in-domain pretraining.
This transfer is possible because LAID tokens depend only on how identities appear within a sequence, so different datasets share the same input space even when their identity vocabularies do not overlap.
We use the three commerce datasets, whose events share a similar user--item structure, take each as the target in turn, and pretrain the pattern encoder on the target alone or on the other two datasets before the same fine-tuning on the target.
As shown in Table~\ref{tab:transfer}, cross-dataset pretraining matches or slightly exceeds in-domain pretraining on all three targets, and both outperform training without pretraining by 3.0--6.0\%.
The identity behavior patterns learned on some platforms are therefore reusable on others, which allows a single pattern encoder to serve multiple scenarios.

\begin{table}[t]
\centering
\caption{PR-AUC with the pattern encoder pretrained on the target dataset alone or on the other two commerce datasets.}
\label{tab:transfer}
\small
\setlength{\tabcolsep}{4pt}
\begin{tabular}{@{}lccc@{}}
\toprule
Pretraining data & Taobao & RetailRocket & Synerise \\
\midrule
Without pretraining & .2608\scriptsize{$\pm$.0042} & .8011\scriptsize{$\pm$.0055} & .4561\scriptsize{$\pm$.0036} \\
Target only         & .2686\scriptsize{$\pm$.0022} & .8283\scriptsize{$\pm$.0067} & .4776\scriptsize{$\pm$.0047} \\
\rowcolor{ptbg}
Other two datasets  & \textbf{.2698}\scriptsize{$\pm$.0025} & \textbf{.8313}\scriptsize{$\pm$.0016} & \textbf{.4836}\scriptsize{$\pm$.0010} \\
\bottomrule
\end{tabular}
\end{table}

\subsection{Efficiency Analysis}
Since identity vocabularies keep growing in online payment systems, an identity representation must remain affordable at scale. To evaluate whether LAID meets this, we compare it with ID embeddings and hashing in representation size and inference time.

\begin{figure}[t]
\centering
\includegraphics[width=0.98\linewidth]{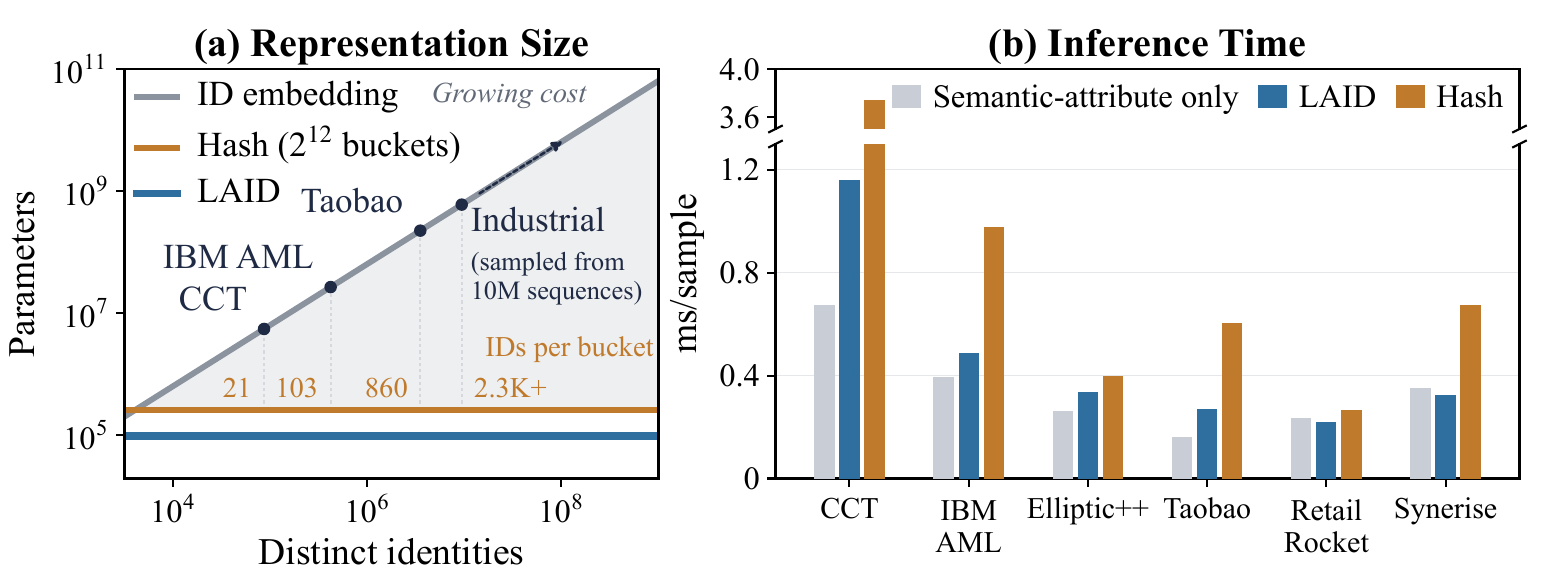}
\caption{Efficiency of identity representations. (a) Parameters per identity field; numbers give identities per hash bucket. (b) Inference time per sample.}
\vspace{-4mm}
\label{fig:efficiency}
\end{figure}

\para{Representation size}
LAID is the only representation whose size does not grow with the number of identities and that never merges distinct identities within a sequence.
As shown in Figure~\ref{fig:efficiency}(a), per-ID embeddings grow linearly with the number of identities, while hashing fixes the size by packing hundreds of identities into each bucket on large datasets.
The LAID vocabulary is bounded by the window length, so its size stays constant as new identities emerge.

\para{Inference time}
LAID is also cheaper to compute than hashing.
As shown in Figure~\ref{fig:efficiency}(b), including the time for identity encoding, LAID is faster than hashing on all datasets, by up to 3.2$\times$ on CCT, and adds little cost over the semantic-attribute-only model.
This is because LAID only checks whether identity values are equal within a window, without mapping each value into a global code space.

\section{Conclusion}

In this paper, we propose \method, a behavioral pretraining framework that learns reusable patterns from high-cardinality identities without relying on their specific values or semantics.
Specifically, the Local Anonymous Identity Tokenizer preserves identity behavior structures through anonymous encoding, while the Multi-Scale Pattern Codebook abstracts these structures into discrete behavior patterns via self-supervised pretraining.
Extensive experiments on industrial payment-risk data and six public datasets demonstrate the effectiveness of \method across financial-risk and commerce tasks.
Our findings highlight identity behavior patterns as a promising source of reusable knowledge for transaction foundation models.

\bibliographystyle{ACM-Reference-Format}
\bibliography{references}
\appendix

\section*{Ethical Considerations}
All identity values in the proprietary dataset were anonymized before modeling, and we report only relative improvements on it to protect commercially sensitive information.
The online evaluation was conducted in shadow mode, so model outputs did not affect any real user or transaction.
Since LAID represents identities by how they recur rather than by their values, it also reduces the reliance of downstream models on specific personal identifiers.

\begin{table}[t]
\centering
\caption{Comparison of strategies for encoding high-cardinality identity fields.}
\label{tab:encoding}
\small
\setlength{\tabcolsep}{5pt}
\begin{tabular}{@{}lccc@{}}
\toprule
Strategy
& \begin{tabular}[c]{@{}c@{}}Size grows\\with IDs\end{tabular}
& \begin{tabular}[c]{@{}c@{}}Unseen\\IDs\end{tabular}
& \begin{tabular}[c]{@{}c@{}}Identity\\behavior\end{tabular} \\
\midrule
Vocabulary-based  & Linear & OOV vector     & Value-tied   \\
Hashing & No     & Arbitrary code & Value-tied   \\
Hand-crafted   & No     & Supported      & Predefined   \\
Semantics-based           & No     & If content     & Not captured \\
\midrule
\rowcolor{ptbg}
LAID (ours)               & No     & Invariant      & Learned      \\
\bottomrule
\end{tabular}
\end{table}

\section{Comparison of Identity Encoding Strategies}
\label{app:encoding}

Identity fields differ from ordinary categorical features in two ways: their vocabularies grow without bound as new cards, devices, and accounts appear, and their raw values carry little semantic meaning.
Table~\ref{tab:encoding} compares existing strategies along three properties that follow from these differences: whether model size grows with the number of identities, how unseen identities are handled, and whether identity behavior is represented.
LAID is the only strategy that keeps a fixed size, treats unseen identities in the same way as seen ones, and learns identity behavior from data.

\para{Vocabulary-based encoding}
Vocabulary-based encoding assigns each identity value its own embedding, as in ID-based sequential models and pretrained models over blockchain addresses~\cite{sun2019bert4rec, zhai2024actions, hu2023bert4eth}.
The embedding table grows linearly with the number of identities, and identities absent from training share a single out-of-vocabulary vector.
Because each embedding is tied to one value, what the model learns about one card or device does not transfer to another that behaves the same way.

\para{Hashing}
Hashing maps identity values into a fixed number of buckets, which bounds memory regardless of vocabulary size~\cite{weinberger2009feature}.
Hash embeddings, compositional embeddings, deep hash embeddings, and frequency-aware compression further reduce collisions or memory~\cite{svenstrup2017hash, shi2020compositional, kang2021learning, hu2024zipzap}.
An unseen identity, however, receives an arbitrary code determined by its hash value, and distinct identities may collide into the same code.
The learned embeddings thus remain tied to values, and collisions can merge identities that a risk model needs to keep apart.

\para{Hand-crafted statistics}
Velocity features and other hand-crafted statistics, such as the number of distinct devices linked to a card within a time window, describe identity behavior independently of raw values and therefore apply to unseen identities.
Each statistic, however, is predefined by domain experts, and covering a new behavior requires designing a new feature.
LAID can be viewed as a learnable generalization of these statistics: its channels record when and how often each identity appears, and the model learns which combinations matter.

\para{Semantics-based encoding}
Semantics-based encoding derives representations from item content, either by quantizing content embeddings into semantic IDs or by encoding descriptions with pretrained text encoders~\cite{rajput2023recommender, singh2024better, hou2023learning, yuan2023where}.
These representations generalize to unseen items that come with descriptive content.
Identity fields such as card numbers, device fingerprints, and IP addresses have no such content, and even when a value can be embedded, its embedding says nothing about how the identity behaves within a sequence.

\para{LAID}
LAID carries the value-independent view of anonymous walks and related graph encodings~\cite{ivanov2018anonymous, wang2021inductive, li2020distance} over to transaction sequences.
It represents each identity by how it appears and recurs within the sequence, so its vocabulary is bounded by the sequence length, its tokens are invariant to relabeling, and it requires no identity content.
Within a sequence, LAID never merges distinct identities, which preserves the sharing structure that hashing may corrupt, and the pattern codebook abstracts this structure into reusable behavior patterns that can be pretrained without labels.
Knowledge about specific identities, such as known fraudulent devices, is typically maintained by rule-based blocklists in deployed risk systems; LAID instead covers what such lists cannot, namely the behavior of identities that are new or not yet flagged.

\section{Dataset Details}
\label{app:datasets}

\subsection{Public Datasets}
\label{app:public}

\para{Datasets and prediction tasks}
We evaluate \method on six public datasets spanning two task families: financial risk detection (CCT, IBM AML, and Elliptic++) and commerce behavior prediction (Taobao, RetailRocket, and Synerise). The financial datasets identify fraudulent, money-laundering, or illicit transactions, whereas the commerce datasets focus on purchase-related prediction. Table~\ref{tab:app_datasets} summarizes the positive class and identity fields for each dataset. Since public datasets rarely release opaque identifiers such as card or device IDs, we use their high-cardinality entity-like fields as identity fields.
Some of these fields, such as city or category, carry partial semantics, but LAID uses only their recurrence structure, so the results reflect the value of identity behavior alone. For Synerise, which lacks event-level labels, we construct a seven-day conversion task from the RecSys 2025 event logs: an add-to-cart event is labeled positive if the same user purchases the same SKU within the following seven days. Targets whose conversion windows extend beyond their respective data splits are excluded.

\begin{table*}[t]
\centering
\caption{Details of the public datasets: positive class, identity fields, number of sequence subjects and supervised targets in each split, and maximum history length $L_{\max}$.}
\label{tab:app_datasets}
\small
\begin{tabular}{@{}lllrrrrr@{}}
\toprule
Dataset & Positive class & Identity fields & Subjects & Train & Valid & Test & $L_{\max}$ \\
\midrule
CCT          & fraud            & merchant name, city     & 6,139     & 125,304   & 3,657,794  & 3,656,690  & 512 \\
IBM AML      & laundering       & receiving bank, account & 496,999   & 15,936    & 760,071    & 760,345    & 512 \\
Elliptic++   & illicit          & two output addresses    & 46,564    & 18,360    & 7,765      & 11,184     & 256 \\
Taobao       & purchase         & item, category          & 987,991   & 2,683,554 & 10,944,062 & 25,628,669 & 100 \\
RetailRocket & purchase         & item                    & 58,653    & 428,761   & 55,209     & 56,467     & 100 \\
Synerise     & 7-day conversion & SKU, URL                & 2,032,633 & 4,944,430 & 529,289    & 642,336    & 512 \\
\bottomrule
\end{tabular}
\end{table*}

\para{Sequence construction}
For each target event, we construct a chronological sequence from the preceding events of the same subject (e.g., a user, card, or account), truncated to the maximum history length $L_{\max}$ given in Table~\ref{tab:app_datasets}. The current event is appended as the prediction target, with its label excluded from the model input; for the commerce tasks, its behavior type is additionally masked. Since Elliptic++ provides no subject-level sequences, we construct each target sequence by tracing the fund flow backward to obtain up to 256 ancestor transactions, with the target transaction itself serving as the sequence subject.

\para{Temporal splits}
All datasets are split chronologically into training, validation,
and test periods rather than by disjoint entities.
A validation or test target may therefore use earlier events of the
same subject, including those from preceding periods, but never
their labels.
Split boundaries follow the native time granularity of each dataset.
CCT and IBM AML use approximately 70\%/15\%/15\% of the global
event timeline, while Synerise is split at 80\% and 90\% of its
time span.
Elliptic++ uses time steps 1--31, 32--39, and 40--49 for training,
validation, and testing, respectively.
Taobao uses 2017-12-01 and 2017-12-02 as split boundaries, while
RetailRocket uses 2015-08-15 and 2015-09-01.

\para{Preprocessing and sampling}
The baselines in Table~\ref{tab:public} do not use identity fields, whereas \method encodes them with LAID as sequence-local identifiers; the hash-augmented baselines in the relabeling study are described in Appendix~\ref{app:relabel}. For the financial datasets, all positive training targets are retained and paired with five randomly sampled negatives per positive. To keep Taobao training tractable, its training targets are subsampled at a rate of 1/20 using a deterministic, label-independent hash. Validation and test sets preserve the natural label distributions, and all reported metrics are computed on these sets.

\subsection{Industrial Dataset}
\label{app:industrial}

\para{Data and sequence construction}
The industrial dataset contains over 100M transaction sequences sampled from a real-world online payment platform, whose full traffic is considerably larger.
Each transaction includes semantic attributes, numerical features such as the amount, and nine high-cardinality identity fields: card, device, seller, client IP address, billing email, account email, shipping phone number, shipping contact name, and shipping postal code.
Sequences are constructed at the user level in chronological order, with each target transaction appended to its preceding history and a maximum sequence length of 300.
All identity values are anonymized before modeling, and due to confidentiality, we report only relative improvements over the reference model.

\para{Offline evaluation}
Transactions from March to June are split chronologically into training, validation, and test periods, with the test period strictly following those used for training and model selection.
Positive labels correspond to confirmed fraudulent transactions.
Besides AUC-ROC, we report two business metrics.
T1 is the recall among the top 1\% highest-scored transactions.
AddCB measures the additional chargeback recall among transactions not rejected or challenged by the existing risk system, under a fixed intervention budget.
Both metrics focus on the high-risk region where operational decisions are made.

\para{Online shadow evaluation}
We further evaluate all variants in shadow mode on sampled production traffic from July to August, following the offline period.
All variants score the same transactions in parallel with the production model.
Since risk labels take several weeks to mature, we compute the metrics after the labels become available, using the same transactions and labels for all variants.
AUC is computed daily and normalized by the average AUC of the semantic-attribute-only model over the first three days.
This enables a paired comparison on production traffic without affecting actual risk decisions.

\section{Implementation Details}
\label{app:impl}

\begin{algorithm}[t]
\caption{Training procedure of \method.}
\label{alg:id2pattern}
\begin{algorithmic}[1]
\Require unlabeled sequences $\mathcal{D}_u$, labeled sequences $\mathcal{D}_l$, window lengths $\{L_s\}_{s\in\mathcal{S}}$
\Statex \textit{Stage 1: self-supervised pattern pretraining}
\For{each sequence $\mathcal{X}\in\mathcal{D}_u$}
  \State $R_s\gets\operatorname{LAID}(X_s)$ for each $s\in\mathcal{S}$ \Comment{single pass with hash maps}
  \State Sample $\mathcal{F}_{\mathrm{mask}}\subseteq\mathcal{F}_{\mathrm{id}}$ and mask its tokens at $\mathbf{x}_T$ in every $R_s$
  \State $\mathbf{h}_s\gets\operatorname{Encoder}(R_s)$, $\mathbf{q}_s\gets$ nearest code in $\mathcal{C}^{(s)}$ \Comment{Eq.~\eqref{eq:code}}
  \State Update pattern branch and recovery heads with $\mathcal{L}_{\mathrm{pre}}$ \Comment{Eqs.~\eqref{eq:recover}--\eqref{eq:vq}}
\EndFor
\Statex \textit{Stage 2: pattern-augmented fine-tuning}
\State Initialize the pattern branch from Stage 1 and discard recovery heads
\For{each labeled sequence $(\mathcal{X},y_T)\in\mathcal{D}_l$}
  \State Compute $\mathbf{Q}=[\mathbf{q}_{\mathrm{S}},\mathbf{q}_{\mathrm{M}},\mathbf{q}_{\mathrm{L}}]$ as in Stage 1, without masking
  \State Build event tokens $\mathbf{v}_1,\ldots,\mathbf{v}_T$ with LAID over $\mathcal{X}$ \Comment{Eq.~\eqref{eq:event}}
  \State $\hat{y}_T\gets\operatorname{Head}\big(\operatorname{Readout}(\operatorname{Backbone}([\mathbf{Q};\mathbf{v}_1,\ldots,\mathbf{v}_T]))\big)$
  \State Update all parameters with $\operatorname{BCE}(\hat{y}_T,y_T)$
\EndFor
\end{algorithmic}
\end{algorithm}

\para{Model configuration}
We use three windows of lengths $L_S = 7$, $L_M = 30$, and $L_L = L_{\max}$, where $L_{\max}$ is dataset-specific sequence length, and each scale-specific codebook contains $K = 64$ codes.

\para{Pretraining}
The recovery losses in Eq.~(\ref{eq:recover}) use $\lambda_a = \lambda_n = \lambda_d = 1$, with occurrence and recurrence targets transformed by $\log(1+x)$ and normalized.
The codebook objective uses $\lambda_{\mathrm{VQ}} = 0.05$ and $\beta = 0.25$, together with a usage-balancing regularizer with weight $\gamma = 0.1$ that encourages all codes to be used and a routing temperature $\tau = 0.2$ that controls the soft code assignment during training.

\para{Optimization}
We use AdamW for both pretraining and fine-tuning with a learning rate of $3\times10^{-4}$ and weight decay of $10^{-4}$, and fine-tune for 100 epochs with early stopping on the validation set.
Hyperparameters are selected on the validation set, and all experiments are conducted on NVIDIA H200 GPUs.

\section{Baseline Details}
\label{app:baselines}

\subsection{Industrial Baselines}
The reference model is the platform's Transformer-based production transaction foundation model, which models transaction histories using semantic attributes and numerical features.
To isolate the effect of identity representation, two variants additionally incorporate identity fields through hashing or LAID, and \method further adds the pretrained multi-scale pattern tokens.
For hashing, each identity value is prefixed with its field name and hashed once with 128-bit MurmurHash3; four positions extracted from the hash each select one of 32 slots, yielding a 128-dimensional multi-hot vector that is fed to a field-specific MLP.
This encoding needs no vocabulary but distinguishes at most $32^4 = 2^{20}$ codes, so distinct identities may collide.
The relabeled variants are evaluated without retraining following Appendix~\ref{app:relabel}.

\subsection{Public Baselines}
We compare against fourteen baselines from three modeling paradigms, following the grouping in Table~\ref{tab:public}.
Baselines marked with $^\dagger$ are reimplemented following their papers, as no official code is available; all others are built on publicly available implementations.

\para{Sequential models}
\begin{itemize}
    \item \textbf{LSTM.} A recurrent encoder with gated memory cells that encodes the event history and is trained end to end with the classification objective.
    \item \textbf{GRU.} A recurrent encoder with simpler gating than LSTM, trained with the same inputs and objective.
    \item \textbf{Transformer.} A self-attention encoder with positional encodings over the event history, serving as the basic attention-based baseline.
\end{itemize}

\para{Tabular and boosting models}
\begin{itemize}
    \item \textbf{XGBoost.} A gradient-boosted tree model on the fields of the current event and aggregated statistics of its history.
    \item \textbf{CatBoost.} A gradient-boosted tree model with native handling of categorical fields, using same features as XGBoost.
    \item \textbf{Temporal Boosting.$^\dagger$} A gradient-boosted tree model that alternates leaf refreshing and tree growth over chronological training shards to follow temporal shifts in the data.
\end{itemize}

\para{Pretraining-based and specialized models}
\begin{itemize}
    \item \textbf{BERT4Rec.} A bidirectional Transformer originally proposed for sequential recommendation; we pretrain it by reconstructing masked historical events and then fine-tune it for classification.
    \item \textbf{TabBERT.} A hierarchical Transformer for tabular event sequences that encodes fields within each event and events within each sequence.
    \item \textbf{HSTU.} A generative sequential model for user behavior that combines gated attention with relative position bias.
    \item \textbf{PANTHER.} A generative pretraining framework for sequential user behavior that pretrains on behavior sequences and is fine-tuned for payment risk detection.
    \item \textbf{IBM AI-on-Z GRU.} A GRU architecture from IBM's open-source fraud-detection project that classifies each transaction from a short history of six events.
    \item \textbf{NVIDIA TFM.} A transaction foundation model~\cite{wu2026transactionfm} that tokenizes each event into field tokens and pretrains a causal decoder with next-token prediction.
    \item \textbf{FRiskGPT.$^\dagger$} A generative financial risk model that treats each event as a behavior token and pretrains a causal decoder with next-behavior prediction.
    \item \textbf{PRAGMA.} A transaction foundation model that pretrains on users' financial event histories and is fine-tuned for downstream risk tasks.
\end{itemize}

\para{Implementation}
All baselines are trained on our datasets with the same splits, as no public checkpoints are available for these tasks.
Each baseline is adapted to the binary classification of the current event, retains its original architecture and history length, and uses only training-period histories for pretraining.
Neural baselines are averaged over three seeds.

\section{Identity Relabeling Protocol}
\label{app:relabel}

For the experiments in Section~\ref{sec:rq3}, we train GRU, Transformer, TabBERT, FRiskGPT, and HSTU with hashed identity fields as additional inputs.
Identity fields are hashed into multi-hot vectors as in Appendix~\ref{app:baselines}, and all other settings, including training steps, seeds, and history lengths, follow Table~\ref{tab:public}.
We then fix the trained models and test targets and replace a fraction of identity values with values never seen during training.
For each identity field, a random one-to-one mapping relabels 0\%, 25\%, 50\%, 75\%, or 100\% of its distinct values, and the same mapping is applied to both historical and current events.
Event order, numerical features, ordinary categorical features, and labels remain unchanged, so the relabeling changes which values identities take while preserving which events share the same identity.
Metrics are averaged over three model seeds.
The industrial relabeled variants in Table~\ref{tab:industrial} follow the same procedure with all identity values relabeled.

\end{document}